%% file: reid_paper_arxiv.tex
\documentclass[10pt,twocolumn,letterpaper]{article}

\usepackage{wacv}
\usepackage{times}
\usepackage{epsfig}
\usepackage{graphicx}
\usepackage{amsmath}
\usepackage{amssymb}
\usepackage{floatrow}
\usepackage{multirow}
\usepackage{subcaption}
\usepackage{sidecap}
\sidecaptionvpos{figure}{t}
\usepackage[table,xcdraw]{xcolor}
\usepackage{hyperref}
\usepackage{url}
\usepackage{listings}
\wacvfinalcopy 

\def\wacvPaperID{***}

\ifwacvfinal\fi
\begin{document}

\title{Video Person Re-Identification using Learned Clip Similarity Aggregation}

\author{Neeraj Matiyali\\
IIT Kanpur\\
{\tt\small neermat@cse.iitk.ac.in}
\and
Gaurav Sharma\\
NEC Labs America\\
{\tt\small grv@nec-labs.com}
}

\maketitle
\ifwacvfinal\thispagestyle{empty}\fi

\def\x{\textbf{x}}
\def\r{\textbf{r}}
\def\c{\textbf{c}}
\def\X{\textbf{X}}
\def\V{\textbf{V}}
\def\c{\textbf{c}}
\def\s{\textbf{s}}
\def\o{\textbf{0}}
\def\I{\textbf{I}}
\def\f{\textbf{f}}
\def\v{\textbf{v}}
\def\p{\textbf{p}}
\def\w{\textbf{w}}
\def\R{\mathbb{R}}
\def\F{\mathcal{F}}
\def\L{\mathcal{L}}
\def\A{\mathcal{A}}
\newcommand\norm[1]{\left\lVert#1\right\rVert}
\newfloatcommand{capbtabbox}{table}[][\FBwidth]

\input{abstract.tex}

\input{intro.tex}

\input{related.tex}

\input{approach.tex}

\input{experiments.tex}

\input{conclusion.tex}

{\small
\bibliographystyle{ieee}
\bibliography{refs}
}

\newpage
\appendix
\input{appendix.tex}

\end{document}

%% file: abstract.tex
\begin{abstract}
We address the challenging task of video-based person re-identification. Recent works have shown
that splitting the video sequences into clips and then aggregating clip based
similarity is appropriate for the task. We show that using a learned clip similarity aggregation
function allows filtering out hard clip pairs, e.g.\ where the person is not clearly visible, is in
a challenging pose, or where the poses in the two clips are too different to be informative. This
allows the method to focus on clip-pairs which are more informative for the task. We also introduce
the use of 3D CNNs for video-based re-identification and show their effectiveness by performing
equivalent to previous works, which use optical flow in addition to RGB, while using RGB inputs
only. 
We give quantitative results on three challenging public benchmarks and show better or
competitive performance. We also validate our method qualitatively.
\end{abstract}

%% file: intro.tex
\section{Introduction}

Person re-identification is the problem of identifying and matching persons in videos captured from
multiple non-overlapping cameras. It plays an important role in many intelligent video surveillance
systems and is a challenging problem due to the variations in camera viewpoint, person pose and appearance, 
and challenging illumination along with various types and degrees of occlusions.

Visual person re-identification involves matching two images or video sequences (containing persons)
to answer whether the persons in the two videos are the same or not. The general approach for it
includes (a) extraction of features that are discriminative wrt.\ the identity of the persons while
being invariant to changes in pose, viewpoint, and illumination and (b) estimating a distance metric
between the features. The earlier methods for re-identification used handcrafted features in
conjunction with metric learning to perform the task \cite{farenzena2010person, gray2008viewpoint, 
kviatkovsky2013color, ma2012local, zhao2013unsupervised}. These works mainly leveraged intuitions
for the task, while in recent years, the use of deep CNNs has become more common owing
to their superior performance \cite{ahmed2015improved, cheng2016person, ding2015deep,
li2014deepreid, xiao2016learning, xiao2017joint}. 

\begin{figure}
    \centering
    \includegraphics[width=0.9\linewidth]{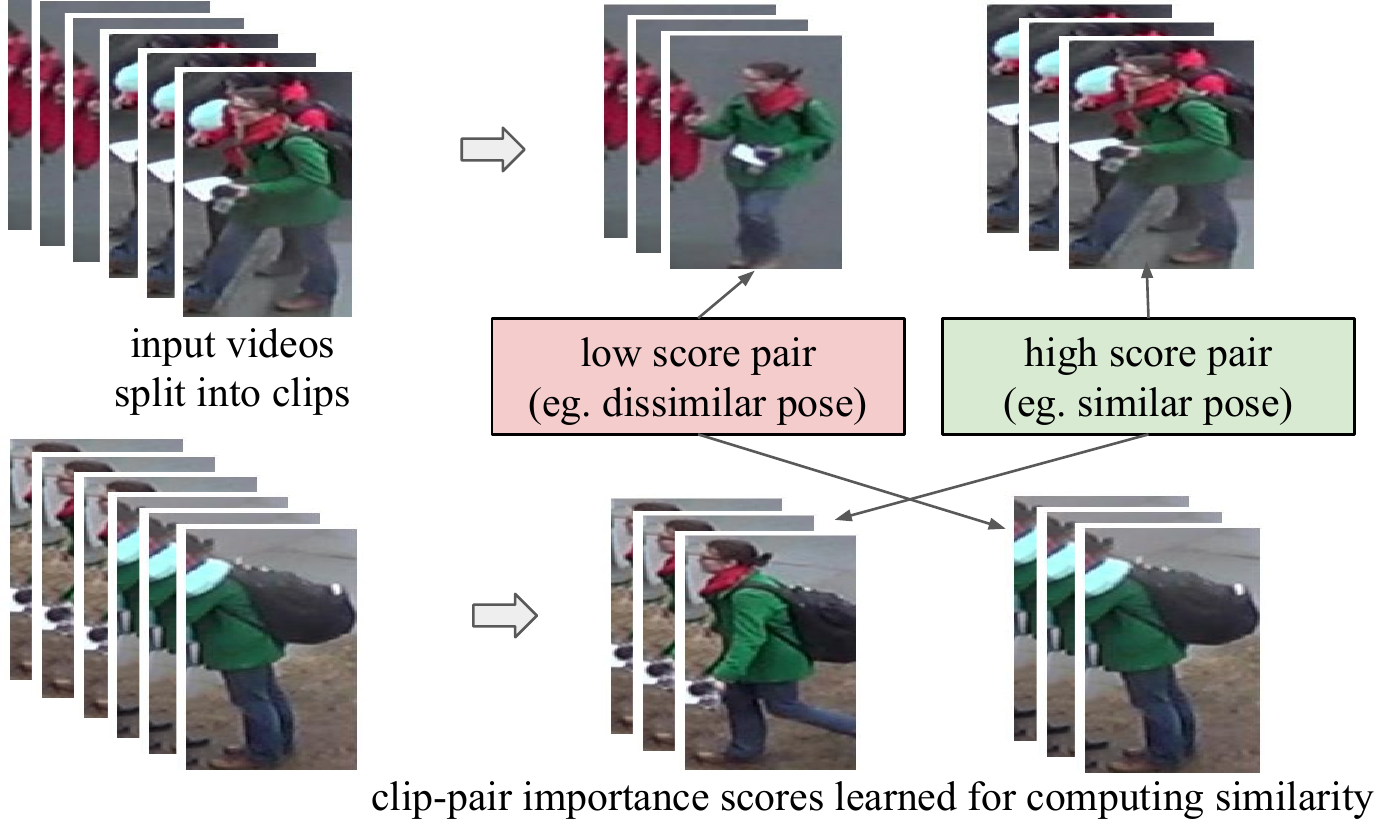}
    \vspace{-0.5em}
    \caption{Illustration of the proposed method. We learn an importance scoring function for
    aggregating clip pairs of video sequences, for person re-identification task. The method learns to
    weight important clip pairs, which help in discrimination, higher than those which are not
    informative. \\
    \vspace{-3em}
    } 
    \label{fig:illus}
\end{figure}

Many of the previous works on person re-identification have focused on image-based benchmarks,
however, with the introduction of large-scale video re-identification benchmarks such as MARS
\cite{zheng2016mars} video-based setting is becoming popular. Most existing methods on video-based
re-identification extract CNN features of individual frames and aggregate them using average
pooling, max pooling, temporal attention mechanisms, or RNNs \cite{mclaughlin2016recurrent,
yan2016person, zheng2016mars, zhou2017see}. These methods, thus, represent the video sequence as a
single feature vector. However, for long sequences that have a significant amount of variation in
pose, illumination, etc., a single vector might not be enough to represent them. 

A recent state-of-the-art video based method by Chen et al.\ \cite{chen2018video} address the
problem by dividing the sequences into short clips, and embedding each clip separately using a CNN
and applying a temporal attention based method. To match two given sequences, they compute similarities
between all pairs of clips, and compute the final similarity by aggregating a fixed percentage of
top clip pair similarities. Thus, the contribution of a clip in a video sequence is dynamically
determined, based on its similarities to the clips in the other sequence. Chen et al.\
\cite{chen2018video} assume that the similarity between a pair of clips is indicative of the
informativeness of the clip pair. We argue that this assumption is not necessarily true in
practice, e.g.\, a pair of clips with low similarity can be utilized as evidence for the fact that
the persons in the two clips are different.  Such clip-pairs get discarded while computing the final
similarity, which may hurt the re-identification performance. Another shortcoming of the method is
that it uses a fixed percentage of the clip-pairs for all pairs of sequences. This limits the
performance of the method since for different pairs of sequences, the number of informative
clip-pairs can vary. 

We address the above shortcomings of Chen et al.\ \cite{chen2018video}, and propose an end-to-end
trainable model to estimate the similarity between two video sequences.  Our model takes pairs of
clips as input in a sequence and predicts an importance score for each clip pair.  It computes the
final similarity between the two sequences by taking an average of the clip-pair similarities
weighted by their corresponding importance scores.  Thus, our model allows filtering of
non-informative or distracting clip-pairs while focusing only on clip-pairs relevant for estimating
the similarities.
While \cite{chen2018video}  aim to filter non-informative or distracting clip-pairs, like
here, the measure of informativeness is different. 
\cite{chen2018video} uses clip-level similarity as a proxy for the
informativeness, while our method uses a learnable scoring function optimized for
the task at the video level. Consider a clip-pair without any
artefact, but with a low clip-similarity due to different
persons being present. While \cite{chen2018video} would reject such a pair 
despite it being informative,
our scoring function would give it high importance to maintain a low overall similarity.

As another contribution, we show effectiveness of 3D CNNs \cite{tran2015learning, carreira2017quo} for obtaining clip features.
3D CNNs, which have been used for various video based tasks such as action recognition in recent years, remain largely unexplored for the task of video based person re-identification.
We show their effectiveness on this task, by reporting performances equivalent to previous works which use optical flow in addition to RGB, while using RGB
inputs only.

We give quantitative results on three video-based person re-identification
benchmarks, MARS \cite{zheng2016mars}, DukeMTMC-VideoReID \cite{ristani2016MTMC, wu2018exploit} and
PRID2011 \cite{hirzer11}. We show that our trainable similarity estimation model performs
better than the top clip-similarity aggregation proposed by Chen et al. \cite{chen2018video}. To
simulate more challenging situations, we also report experiments with partial frame corruption,
which could happen due to motion blur or occlusions, and show that our method degrades gracefully
and performs better than the competitive baseline. We also provide qualitative results that verify
the intuition of the method.

%% file: related.tex
\section{Related Work}

\paragraph{Image based Re-Identification.}
Initial works on person re-identification focused on designing and extracting discriminative
features from the images \cite{gray2008viewpoint, farenzena2010person, ma2012local,
kviatkovsky2013color, zhao2013unsupervised}. These works mainly leveraged intuitions for the task
and proposed hand designed descriptors that capture the shape, appearance, texture, and other
visual aspects of the person. Other works proposed better metric learning methods for the task of
person reidentification \cite{gray2008viewpoint, prosser2010person, zheng2011person,
koestinger2012large, liao2015efficient, paisitkriangkrai2015learning}. This line of work mainly
worked with standard features and innovated on the type and better applicability of metric learning
algorithms for the task.  

More recent methods, have started leveraging CNN features for the task of person re-identification.
These methods explore various CNN architectures and loss functions. Li et al.~\cite{li2014deepreid}
proposed a CNN architecture specifically for the re-identification task, which was trained using a
binary verification loss. Ding et al.~\cite{ding2015deep} proposed a triplet loss to learn CNN
features. Ahmed et al.~\cite{ahmed2015improved} proposed a siamese CNN architecture and
used binary verification loss for training. Cheng et al.~\cite{cheng2016person} used a parts-based CNN
model for re-identification, which was learned using a triplet loss. Xiao et
al.~\cite{xiao2016learning} used domain guided dropout that allowed learning of CNN features from
multiple domains. They used a softmax classification loss to train the model. Xiao et
al.~\cite{xiao2017joint} jointly trained a CNN for pedestrian detection and identification. They
proposed online instance matching (OIM) loss, which they showed to be more efficient than the
softmax classification loss.

Another line of work \cite{su2017pose, zheng2017pose, zhao2017spindle, suh2018part} leverages human
pose estimators and uses parts-based representations for person re-identification. For example, Suh
et al.~\cite{suh2018part} used a two-stream framework with an appearance and a pose stream, which were
combined using bilinear pooling to get a part-aligned representation.
\vspace{0.5em} \ \\ 
\textbf{Video-based person re-identification.} The methods working with videos commonly rely on
CNNs to extract features from the individual frames, while using different ways for aggregating
frame-wise CNN features, e.g.\ Yan et al.~\cite{yan2016person} used LSTM to aggregate the frame-wise
features. Zheng et al.~\cite{zheng2016mars} aggregated the CNN features using max/average pooling,
and also used metric learning schemes such as KISSME \cite{koestinger2012large} and XQDA
\cite{liao2015efficient} to improve the re-identification performance. McLaughlin et al.
\cite{mclaughlin2016recurrent} used RNN on top of CNN features followed by temporal max/average
pooling.

More recent works have also started exploring temporal and spatial attention based methods for
video-based re-identification. Zhou et al.~\cite{zhou2017see} used a temporal attention mechanism for
weighted aggregation of frame features. Li et al.~\cite{li2018diversity} employed multiple spatial
attention units for discovering latent visual concepts that are discriminative for
re-identification. They combined the spatially gated frame-wise features from each spatial attention
unit using temporal attention mechanisms and concatenation. 

Liu et al.~\cite{liu2017video} used the two-stream framework for video re-identification, which
consists of an appearance and a motion stream, to exploit the motion information in the video
sequences. Instead of using pre-computed optical flow, however, they learned the motion context from
RGB images in an end-to-end manner.

%% file: approach.tex
\section{Approach}\label{sec:approach}

We assume humans have been detected and tracked and we are provided with cropped videos which contain a single human. 
We view the videos as ordered sequence of tensors (RGB frames).
We formally define the problem we address as that of learning a parameterized similarity between two ordered
sequences of tensors. Denote the query and the gallery video sequence as, $\X_q = \{\x_{q,1},
\x_{q,2},\ldots,\x_{q,n}\},$ and $\X_g=\{\x_{g,1}, \x_{g,2},\ldots,\x_{g,m}\}$, with $\x_{i,k} \in
\F = \R^{3\times H \times W}$ being an RGB frame. We are interested in learning a function
$\psi_\Theta: \F^n \times \F^m \rightarrow \R$, with parameters $\Theta$, which takes as input two
sequences, $\X_q, \X_g$ and outputs a real valued similarity between them $\psi_\Theta(\X_q, \X_g)$,
where a high (low) similarity indicates that they are (not) of the same person.

\subsection{Learning Clip Similarity Aggregation}

\begin{figure*}[t]
\centering
    \includegraphics[width=\textwidth]{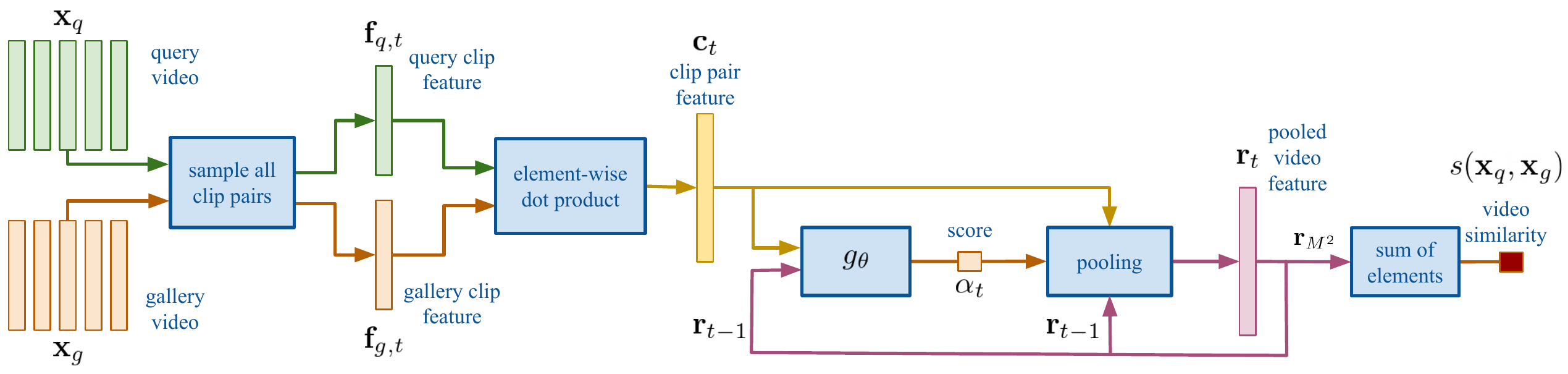}
    \vspace{-2em}
    \caption{The block diagram of the proposed video similarity estimation method. The video sequences are
    first split into clips, which are combined to give a clip features. The combined clip features
    are then pooled with an importance score as a weight. The final pooled representation vector is then
    used to compute the similarity.} \label{fig:SimEst}
\end{figure*}

The similarity function we propose is based on a learned aggregation of clip-pairs sampled from the
video sequences. Fig.~\ref{fig:SimEst} gives a full block diagram of our method. We uniformly sample
$M$ clips of length $L$ from both the query and the gallery sequences, denoted by $\{\s_q^1, \ldots,
\s_q^M\}$ and $\{\s_g^1, \ldots, \s_g^M\}$, where, $\s_q^i, \s_g^i \in \mathbb{R}^{L\times 3 \times
H \times W}$. The number of clips could also be different for the two sequences being compared, but for
brevity and implementation ease we keep them to be the same, allowing potential overlap of the clips
if the number of frames in the sequence(s) is less than $ML$. 

We first forward pass the clips through a state-of-the-art 3D CNN $f_\xi(\cdot)$ with
parameters $\xi$ to obtain $D$-dimensional features $\x_q \rightarrow \{\f_q^1, \ldots, \f_q^M\}$
and $\x_g \rightarrow \{\f_g^1, \ldots, \f_g^M\}$, where, $\f_q^i = f_\xi(\s_q^i)$, $\f_g^i =
f_\xi(\s_g^i)$  and $\f_q^i, \f_g^i \in \mathbb{R}^{D}$. We then learn to estimate which pairs of
clips are informative, considering all the $M^2$ combinations. This is in contrast to many sequence
modeling approaches, like those based on max/average pooling \cite{zheng2016mars} or attention-based
temporal pooling \cite{zhou2017see}, which encode the clip sequences individually with the intuition
that some clips might be bad due to occlusion, difficult pose or high motion blur etc. In our case
we argue that even if some clips have partial artifacts, due to the various nuisance factors, they
might still match with a similarly (partially) corrupted clip from another video, and thus should
not be discarded. Hence, in the proposed method we consider all the quadratic combinations of pairs
of clips and learn to weight them according to their importance. We run the importance estimation
in a sequential manner and condition on the information that we have already accumulated at any step $t$.
We estimate the importance score of the clip pair at step $t$, $\alpha_t$, using a small
neural network $g_\theta(\cdot)$ which takes as input the difference of the aggregated
representation $r_t$ till that point and the combined representation $c_t$ of current clip
pair. The combined representation used for a pair of clips is an element-wise dot product (denoted
as $\odot$) of the clip features, and the pooling process, at step $t = 2,\ldots,M^2$ is given by,
\begin{align}
    \c_t & = \f_{q,t} \odot \f_{g,t}, \; \; \; \; \alpha_{t} = g_\theta ( \r_{t-1} - \c_t) \\
    \r_t & = \frac{1}{\A_t} \left\{\left(\sum_{i=1}^{t-1}\alpha_i \right)\r_{t-1} + \alpha_t \c_t\right\} 
    = \frac{1}{\A_t} \sum_{i=1}^{t}\alpha_i \c_i
\end{align}
with, $\A_t = \sum_{i=1}^{t}\alpha_i, \r_1 = \f_{q,1} \odot \f_{g,1}$. This gives the final
combined representation $\r_{qg} = \r_{M^2}$.

We then predict the similarity score between $\x_q$ and $\x_g$ by taking an average of all clip-pair
cosine similarities weighted according to the importance scores, 
\begin{align}
    s(\x_q, \x_g) = \dfrac{1}{\sum_{t=1}^{M^2}\alpha_t}\sum_{t=1}^{M^2} \alpha_t \dfrac{\f_{q,t}\cdot\f_{g,t}}{\norm{\f_{q,t}}_2\norm{\f_{g,t}}_2}
    \label{eqn:similarity1}
\end{align}
If the clip features $\f_{q,t}$ and $\f_{g,t}$ are $\ell_2$-normalized, then the final similarity \eqref{eqn:similarity1} can be directly computed using the final combined representation $\r_{qg}$ as
\begin{align}
    s(\x_q, \x_g) &= \sum_{l=1}^D r_{qg}^l,
    \label{eqn:similarity2}
\end{align}
where, $\r_{qg} = \left[r_{qg}^1, \ldots, r_{qg}^D\right]$. The expression \eqref{eqn:similarity2}
can be obtained from \eqref{eqn:similarity1}, 
with $\c_{t} = \left[c_{t}^1, \ldots, c_{t}^D\right]$,
as follows, 
{\scriptsize
\begin{align}
    s(\x_q, \x_g) &= \dfrac{1}{\sum_{t=1}^T \alpha_t}\sum_{t=1}^T \alpha_t (\f_{q,t}\cdot\f_{g,t}) 
             = \dfrac{1}{\sum_{t=1}^T \alpha_t}\sum_{t=1}^T \left(\alpha_t \sum_{d=1}^D c_t^d\right) \\
             &= \sum_{d=1}^D \left(\dfrac{1}{\sum_{t=1}^T \alpha_t} \sum_{t=1}^T \alpha_t  c_t^d \right)
              = \sum_{l=1}^D r_{qg}^l.  
              \label{eqn:similarity5}
\end{align}
}

\subsection{Learning}

Our method allows us to learn all the parameters, $\Theta = (\xi, \theta)$ end-to-end and jointly
for the task using standard backpropagation algorithm for neural networks. However, due to
computational constraints, we design the training as a two step process. First, we learn the
parameters of 3D CNNs, then we fix the 3D CNNs and learn the clip-similarity aggregation module
parameters. We now describe each of these steps.
\vspace{0.5em} \ \\ 
\textbf{3D CNN.} In each training iteration, following
\cite{hermans2017defense}, we randomly sample a batch of $PK$ sequences belonging to $P$ person
identities with $K$ sequences from each identity. Then, we randomly sample one clip of length
$L$ frames from each sampled sequence to form the mini-batch. We use a combination of the hard
mining triplet loss \cite{hermans2017defense} and the cross-entropy loss as our
objective,
$    \L(\xi) = \L_\text{triplet}(\xi) + \L_\text{softmax}(\xi).$

The hard mining triplet loss is given as, $\L_\text{triplet}(\xi)=$
{\scriptsize
\begin{multline}
    \sum_{i=1}^P \sum_{a=1}^K
    \Biggl[ 
    m + \max_{p=1,\ldots,K}d(\x_{a,i},\x_{p,i})
    - \min_{\substack{j=1,\ldots,P\\n=1,\ldots,K\\j\ne i}}d(\x_{a,i}, \x_{n,j})
    \Biggr]_+,
    \label{eqn:triplet_loss_i3d}
\end{multline}
}
where, $d(\x_1,\x_2) = \norm{\x_1 - \x_2}_2$, $\x_{k,i}$ is the 3D CNN feature vector of the $k$-the clip of the $i$-th person in the
batch, and $m$ is the margin, and $\left[\cdot\right]_+ = \max(0, \cdot)$. 

We add a classification layer on top of our 3D-CNN network with $C$ classes, where $C$ is the total
number of identities in the training set. Let $\{\w_1, \ldots, \w_C\}$ be the weights of the
classification layer. The softmax cross-entropy loss is  given by,
{\scriptsize
\begin{equation}
    \L_\text{softmax}(\xi) = - \sum_{i=1}^P \sum_{k=1}^K
    \left[ \log \dfrac{\exp(\w_{y^i} \cdot \f_{k,i})}{\sum_{c=1}^C\exp(\w_{c} \cdot \f_{k,i})}
    \right],
\end{equation}
}
where, $y^i$ is the person index of the $i$-th person in the batch. Note that, while learning 3D CNN parameters, $\xi$, we do not use our clip-similarity aggregation module.
\vspace{0.5em} \ \\ 
\textbf{Clip similarity aggregation module.} For learning $\theta$, we use
the same batch sampling process as described for the learning of 3D CNN parameters $\xi$, except now
we uniformly sample $M$ clips of length $L$ instead of a single clip from each sampled sequence. We
extract features $\x_i$ of the clips, with the above learned 3D CNNs, and normalize them.
Then, we compute the similarity scores between all pairs of sequences in the batch using
\eqref{eqn:similarity2}. We use the hard mining triplet loss similar to \eqref{eqn:triplet_loss_i3d}
as the objective, with the euclidean distances replaced by negative clip similarities as defined
above in \eqref{eqn:similarity1}--\eqref{eqn:similarity5}.

%% file: experiments.tex
\section{Experiments and Results}

\subsection{Datasets}
\noindent
\textbf{MARS.} The MARS dataset \cite{zheng2016mars} is a large scale video-based person
re-identification benchmarks.  It contains 20,478 pedestrian sequences belonging to 1261 identities.
The sequences are automatically extracted using DPM pedestrian detector
\cite{felzenszwalb2010object} and GMMCP tracker \cite{dehghan2015gmmcp}. The lengths of the
sequences range from 2 to 920 frames.  The videos are captured from six cameras and each identity is
captured from at least two cameras.  The training set consists of 8,298 sequences from 625
identities while the remaining 12,180 sequences from 636 identities make up the test set which
consists of a query and a gallery set.
\vspace{0.5em} \ \\
\textbf{DukeMTMC-VideoReID.} The DukeMTMC-VideoReID \cite{ristani2016MTMC, wu2018exploit} is another large benchmark of video-based person re-identification. 
It consists of 702 identities for training, 702 identities for testing.
The gallery set contains additional 408 identities as distractors.
There are total 2,196 sequences for training and 2,636 sequences for testing and distraction. Each sequence has 168 frames on average.
\vspace{0.5em} \ \\
\textbf{PRID2011.} PRID2011 dataset \cite{hirzer11} contains 400 sequences of 200 person
identities captured from two cameras.
Each image sequence has a length of 5 to 675 frames.  Following the evaluation protocol from
\cite{wang2014person, zheng2016mars}, we discard sequences shorter than 21 frames and use 178
sequences from the remaining for training and rest 178 sequences for testing.

\subsection{Implementation Details}\label{sec:imp-details}

\begin{figure}
\centering
\includegraphics[height=3.0cm,trim=235 0 0 0, clip]{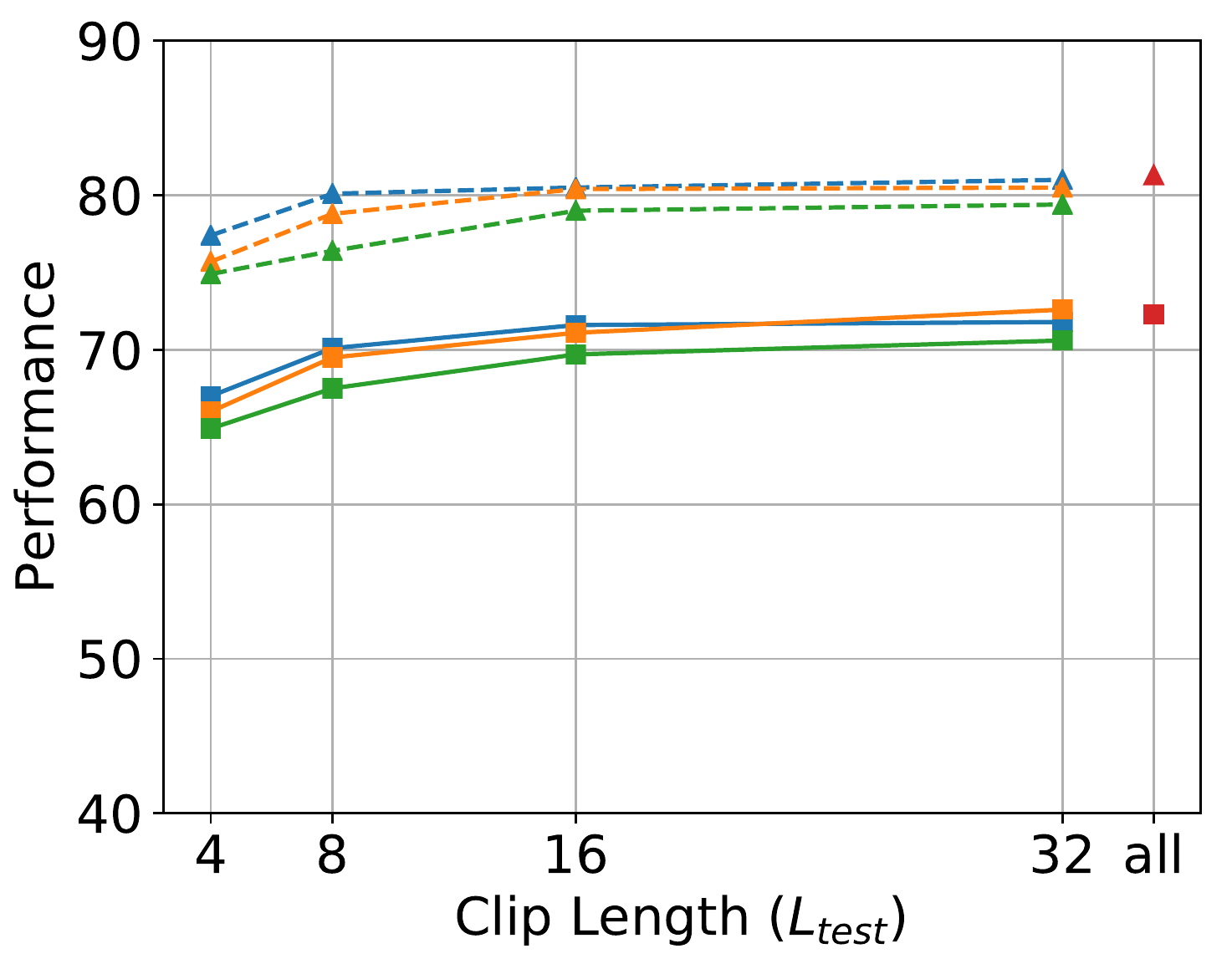} 
\hfill
\includegraphics[height=3.1cm,trim=0 50 440 0, clip]{figs/i3d_run6_eval.pdf} \\
\centering
\includegraphics[height=3.0cm]{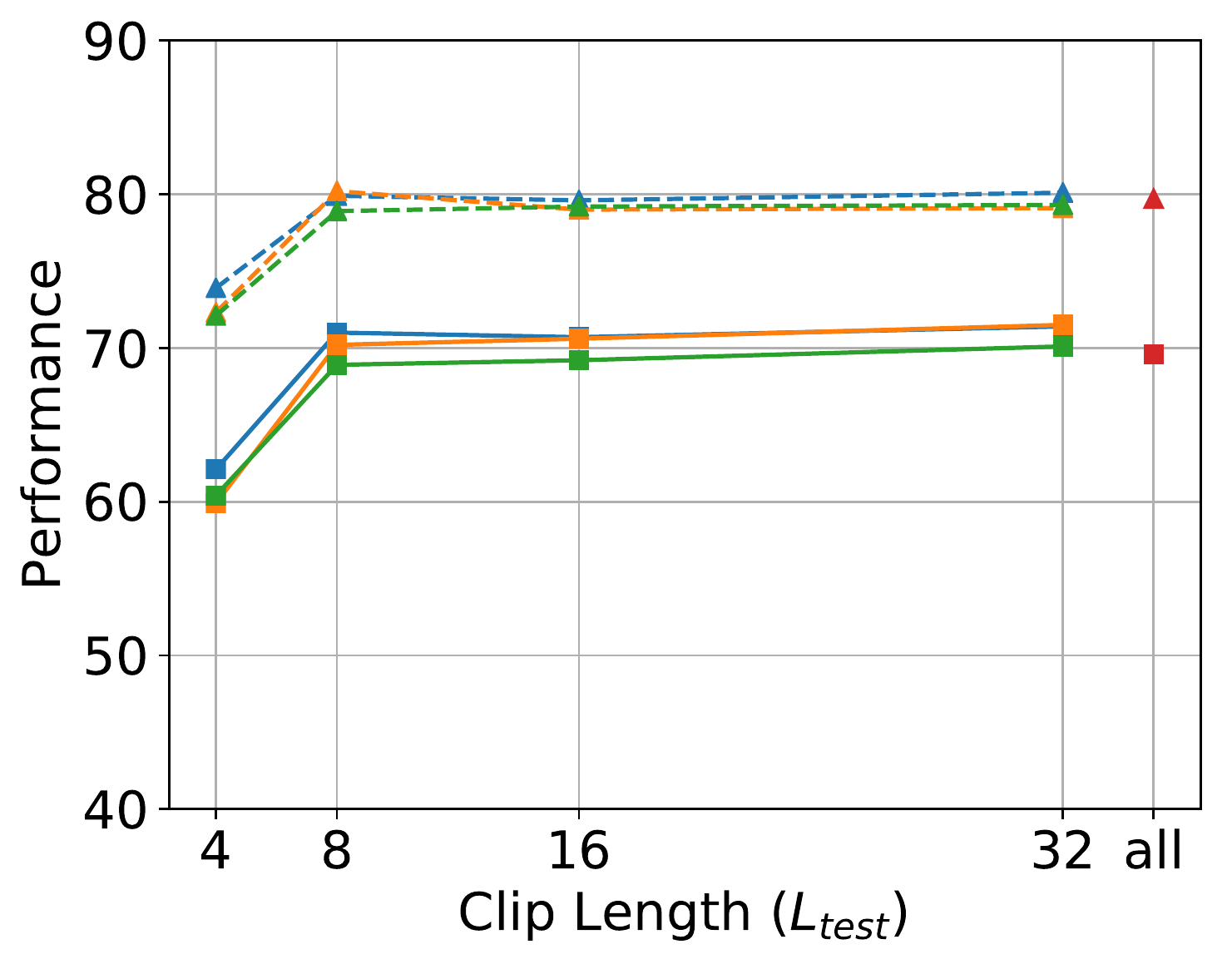}
\hfill
\includegraphics[height=3.0cm]{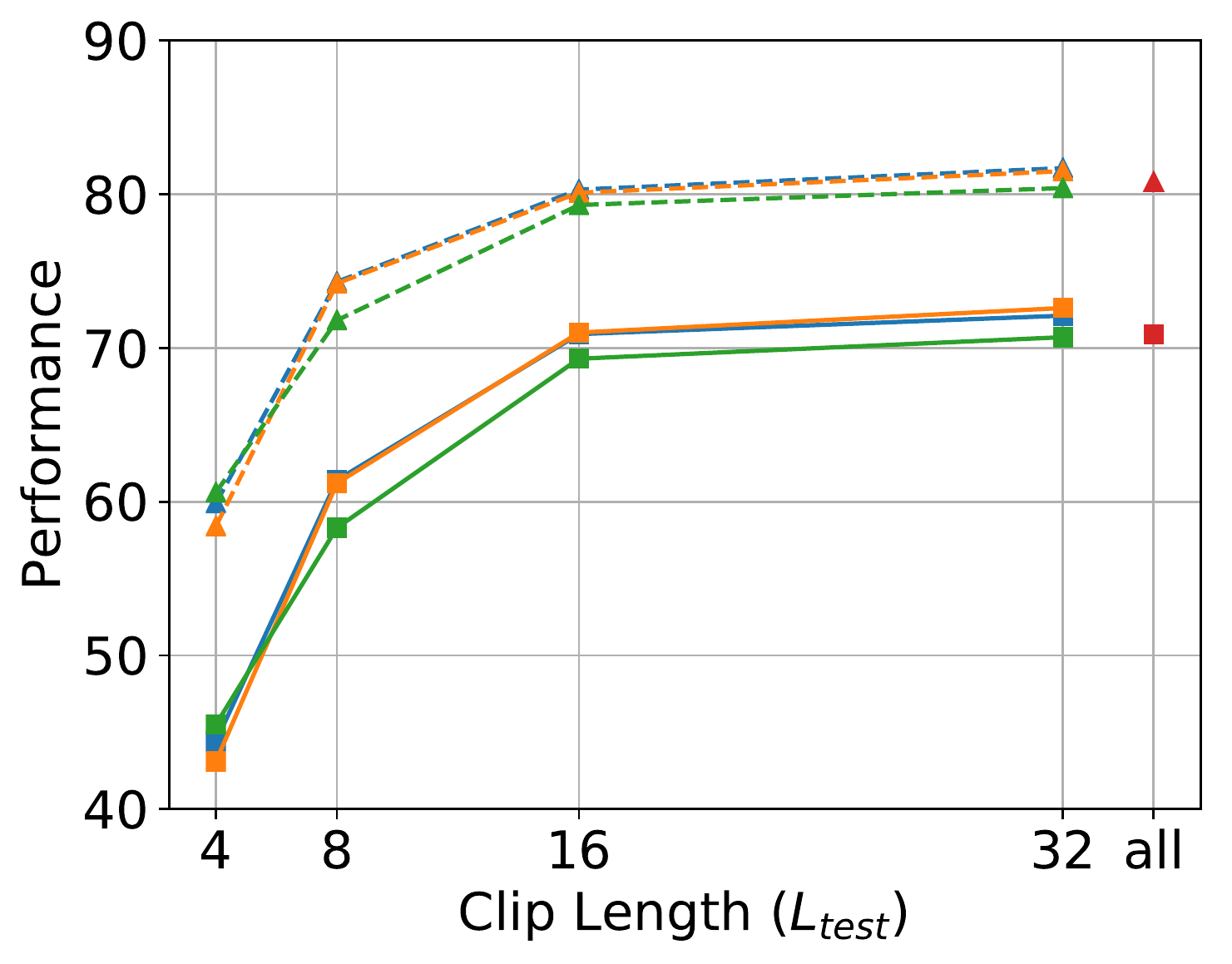}
\vspace{-0.75em}
\caption{Effect of frame sampling methods and test clip length on MARS.
$L_\text{train}=4,8,16$ counter clockwise. 
}
\label{fig:sampling}
\end{figure}

\noindent 
\textbf{3D CNN Architecture.} We use the pytorch implementation\footnote {\url{https://github.com/piergiaj/pytorch-i3d}} of Inception-V1 I3D network \cite{carreira2017quo} pretrained on the Kinetics action recognition dataset.
We remove the final classification layer from the I3D network and replace the original average pooling layer of kernel $2\times 7\times 7$ with a global average pooling layer.
The resulting I3D network takes an input clip of size $L\times 3 \times 256 \times 128$ and outputs a 1024-dimensional feature vector ($D=1024$).
\vspace{0.5em} \ \\
\textbf{Clip similarity aggregation module architecture.} The clip-pair similarity aggregation
module takes as input a pair of tensors $(M\times D, M\times D)$ representing I3D features of $M$
clips sampled from the two sequences to be matched.  In our experiments, we set the number of clips $M$
to 8 and the clip length $L$ to 4 frames, this setting was faster than higher $L$ and smaller $M$
while giving similar performance (kindly see the supplementary document for complete ablation experiment). 
The importance scoring function $g_\theta(\cdot)$ consists of two hidden layers with 1024 units in both layers.
The output layer has a single unit that represents the estimated importance score.
The hidden layers have ReLU activation function while the output layer has the softplus activation function, $\sigma(x) = \log(1 + \exp(x))$.
The softplus function, a smooth approximation of ReLU function, constraints the importance score to always be positive.
We also use a dropout layer \cite{srivastava2014dropout} with dropout probability $0.5$ and a batch normalization layer \cite{ioffe2015batch} after both hidden layers.
\vspace{0.5em} \ \\
\textbf{Training details.} Due to lack of space, we include the complete training details of the 3D
CNN and the Clip Similarity Aggregation module in the supplementary document. 
\vspace{0.5em} \ \\
\textbf{Evaluation protocol and evaluation metrics.} 
We follow the experimental setup of \cite{wang2014person}, \cite{zheng2016mars} and \cite{wu2018exploit} for PRID2011, MARS and DukeMTMC-VideoReID respectively.
For MARS and DukeMTMC-VideoReID, we use the train/test split provided by \cite{zheng2016mars} and \cite{wu2018exploit}, respectively.
For PRID2011, we average the re-identification performance over 10 random train/test splits.
We report the re-identification performance using CMC (cumulative matching characteristics) at selected ranks and mAP (mean average precision).

\subsection{Analysis of I3D Features for Re-Identification}

\begin{SCfigure}
\centering
\includegraphics[width=0.5\linewidth]{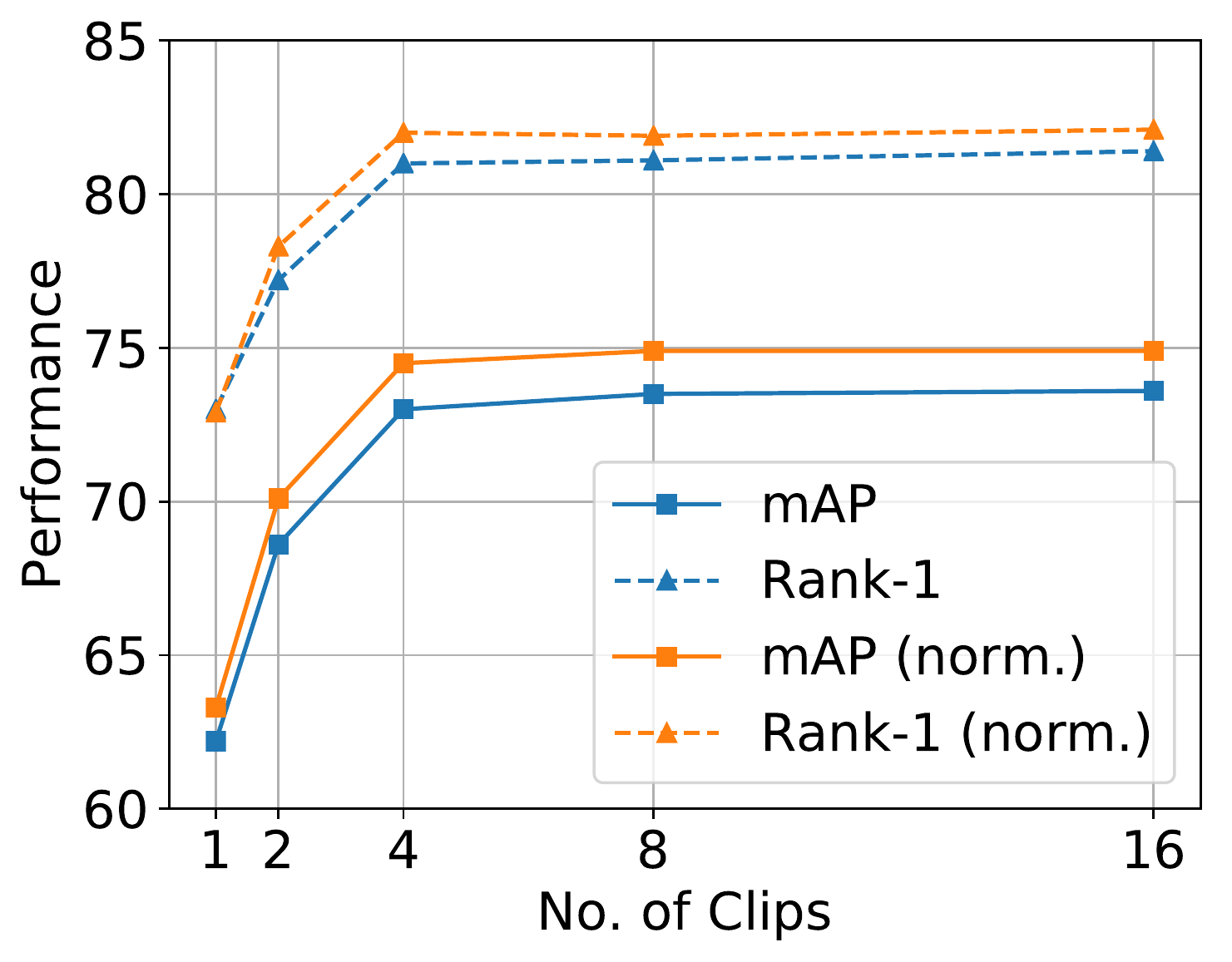}
\caption{Test performance (MARS) with averaging of I3D features of multiple clips, with and without
$\ell_2$-normalization.}
\vspace{-1em}
\label{fig:average-multiple-clips}
\end{SCfigure}

\noindent
\textbf{Frame sampling method and clip length.} In the scenario, where we use a single clip to
represent a sequence, it becomes important how we sample the frames from the sequence to form a
clip. In this experiment, we explore multiple frame sampling methods given in Tab.~\ref{tab:sampling}
and their effect on the re-identification performance. 

\begin{SCtable}
\resizebox{0.68\linewidth}{!}{
\begin{tabular}{l|p{15em}}
\hline
\texttt{consec}& Randomly sample a clip of $L$ consecutive frames \\
\texttt{random}& Randomly sample $L$ frames (arrange in order) \\
\texttt{evenly}& Sample $L$ frames uniformly \\
\texttt{all}   & Take all frames  \\
\hline 
\end{tabular}
}
\vspace{-0.5em}
\caption{Frame sampling methods for clip construction}
\label{tab:sampling}
\end{SCtable}

Note that, all sampling methods in Tab.~\ref{tab:sampling} result in a clip of length $L$ except the \texttt{all}
sampling method. We train three I3D models with different clip lengths, $L_\text{train}\in \{4, 8,
16\}$. The frames are sampled consecutively (\texttt{consec}) to form a clip during training. During
evaluation, each test sequence is represented by I3D features of a single clip sampled in one of the
ways described above. Given a query, the gallery sequences are ranked based on the distances of
their I3D features. We evaluate the three models with different frame sampling methods and test clip
lengths $L_\text{test} \in \{4, 8, 16, 32\}$.

Figure~\ref{fig:sampling} shows the plots of re-identification performance as a function of
$L_\text{test}$ with different frame sampling methods.
We observe that the performance improves as we increase the clip-length during testing, although with diminishing returns.
We also observe that when tested on longer clips (e.g. $L_\text{test}=16, 32$), models trained on
different clip-lengths ($L_\text{train} = 4, 8, 16$) show similar performance to each other.  On the
other hand when tested on shorter clips (e.g. $L_\text{test}=4$), a model trained on shorter clips
performs better than the model trained on longer clips.

The sampling methods \texttt{random} and \texttt{evenly} perform better than the \texttt{consec} sampling method,
especially for smaller clip lengths. 
This can be explained by the fact that \texttt{random} and
\texttt{evenly} have larger temporal extent than \texttt{consec} and do not rely on frames only from
a narrow temporal region which could be non-informative because of difficult pose, occlusion etc.
\vspace{0.5em} \ \\
\textbf{Averaging features of multiple clips.} 
Since sequences in the MARS dataset can be up to
920 frames long, using single short clips to represent these sequences is not optimal. In this
experiment, we take average of I3D features of multiple clips evenly sampled from the original
sequence to represent these sequences. We vary the number of clips in $\{1, 2, 4, 8, 16\}$ on the MARS
dataset. We use the model trained with $L_\text{train} = 4$ and we keep the same clip length
$L_\text{test} = 4$ during the evaluation. We also evaluate with and without the $\ell_2$-normalization of
clip-features. Figure~\ref{fig:average-multiple-clips} shows the test re-identification performance
for different number of clips with and without $\ell_2$-normalization of clip-features. We observe that
averaging features from multiple clips significantly improves the re-id performance. The performance
improves up to around 8 clips beyond which there is little improvement. We also find that
$\ell_2$-normalization of clip features leads to consistent improvement in performance.

\begin{figure}
\includegraphics[width=0.48\textwidth]{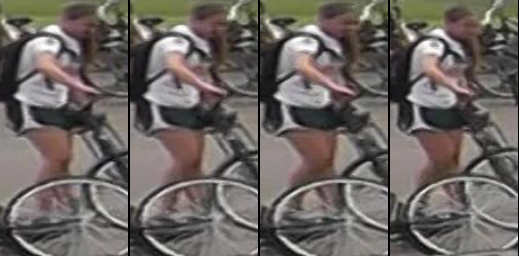}
\hfill
\includegraphics[width=0.48\textwidth]{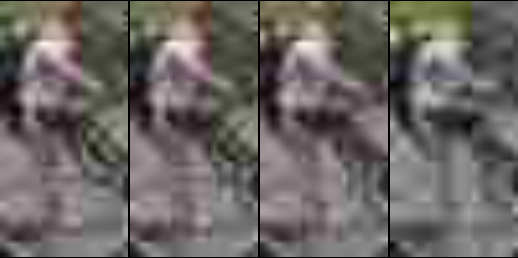}
\vspace{-1em}
\caption{Example of an uncorrupted and a corrupted clip.}
\label{fig:corrupt-uncorrupt-clip}
\end{figure}

\subsection{Evaluation of Learned Clip Similarity Aggregation on MARS}

In this section, we present the re-identification performance results of our learned clip similarity aggregation method on the MARS test set.
We also investigate the robustness of our method by evaluating it with varying degrees of input corruption.
We randomly corrupt clips during training and evaluation as follows.
For every training or test sequence $\x$, we first randomly pick a number $M_c(\x)$ with $0 \le M_c(\x) \le M_{c}^{\max}$.
Here, $M_{c}^{\max}$ denotes the maximum number of corrupt clips in a sequence with $M_{c}^{\max} \le M$.
Next, we apply a corruption transformation function to randomly selected $M_c(\x)$ of the $M$ clips sampled from the sequence $\x$.
The corruption transformation function consists of first scaling down every frame in the clip by a factor of 5, JPEG compression of resulting scaled down frames, and finally rescaling of the frames up to the original size.
Figure~\ref{fig:corrupt-uncorrupt-clip} shows examples of uncorrupted and corrupted clips.

Let $\{\f_q^1, \ldots, \f_q^M\}$ and $\{\f_g^1, \ldots, \f_g^M\}$ be the $\ell_2$-normalized I3D features of $M$ clips sampled from a query sequence $\x_q$ and a gallery sequence $\x_q$ respectively. As described in Section~\ref{sec:approach}, the similarity between $\x_q$ and $\x_g$, as estimated by our method, is given by \eqref{eqn:similarity1} or \eqref{eqn:similarity2}. We train and evalulate our clip-similarity aggregation module for different rates of input corruption. The rate of input corruption is changed via the parameter $M_c^{\max}$. We use the I3D network trained only on uncorrupted clips and keep it fixed throughout the experiment.

\begin{SCfigure}
\centering
\includegraphics[width=0.7\linewidth]{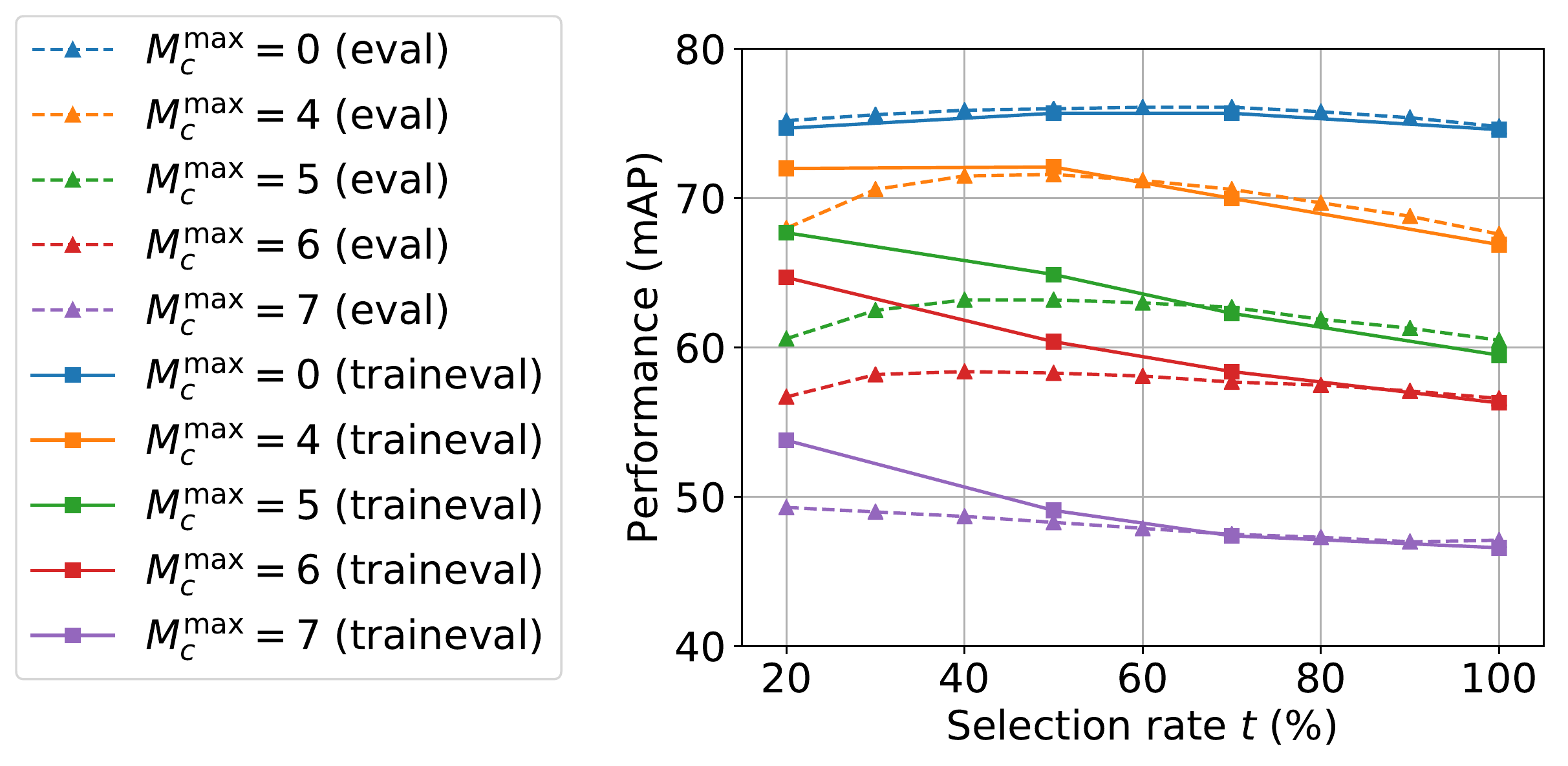}
\vspace{-.8em}
\caption{MARS test mAP vs.\ selection rate ($t$) for top-$t\%$
clip-similarity aggregation.} 
\label{fig:top-t-vs-t}
\end{SCfigure}

\begin{table*}
\centering
\resizebox{0.8\textwidth}{!}{
\begin{tabular}{c|c|llllllll|llllllll}
\hline
 \multirow{3}{*}{Method}                     &     \multirow{3}{*}{ $t$ (\%)}             & \multicolumn{8}{c|}{{mAP}}                                                                                              & \multicolumn{8}{c}{{Rank-1}}                                                                                           \\
                                   \cline{3-18}
           &            & \multicolumn{8}{c|}{$M_c^{\max}$}                                                                                                     & \multicolumn{8}{c}{$M_c^{\max}$}                                                                                                     \\
                                   &                                      & 0             & 1             & 2             & 3             & 4             & 5             & 6             & 7             & 0             & 1             & 2             & 3             & 4             & 5             & 6             & 7             \\
                                   \hline
\multirow{10}{*}{\texttt{topt-e}}     & 10                                   & 73.9          & 72.6          & 71.3          & 68.5          & 63.9          & 57.0            & 53.5          & 49.2          & 78.9          & 78.3          & 78.0          & 76.9          & 73.0          & 67.5          & 64.7          & 62.4          \\
                                   & 20                                   & 75.2          & 74.7          & 73.7          & 72.3          & 68.0            & 60.6          & 56.7          & 49.3          & 80.9          & 80.7          & 80.9          & 80.0          & 76.8          & 70.7          & 67.6          & 62.1          \\
                                   & 30                                   & 75.6          & 75.1          & 74.6          & 73.7          & 70.6          & 62.5          & 58.2          & 49.0            & 81.2          & 81.1          & 81.6          & 81.1          & 79.7          & 73.7          & 69.3          & 61.8          \\
                                   & 40                                   & 75.9          & 75.3          & 75.0            & 74.1          & 71.5          & 63.2          & 58.4          & 48.7          & 81.8          & 81.5          & 82.6          & 81.6          & 80.2          & 73.9          & 70.1          & 61.5          \\
								   & 50                                   & 76.0            & \textbf{75.4}          & 75.1          & 74.3          & 71.6          & 63.2          & 58.3          & 48.3          & 82.5          & 81.5          & 82.6          & 82.2          & \textbf{80.3}          & 73.9          & 70.0          & 61.3          \\
                                   & 60                                   & \textbf{76.1} & \textbf{75.4} & 74.8          & 74.1          & 71.2          & 63.0            & 58.1          & 47.9          & \textbf{83.2} & \textbf{82.3} & 82.2          & 82.2          & 79.7          & 72.9          & 70.1          & 60.7          \\
                                   & 70                                   & \textbf{76.1} & 75.2          & 74.5          & 73.4          & 70.6          & 62.7          & 57.7          & 47.5          & \textbf{83.2} & 82.2          & 82.4          & 81.4          & 79.2          & 72.8          & 69.6          & 60.4          \\
                                   & 80                                   & 75.8          & 75.0            & 73.9          & 72.7          & 69.7          & 61.9          & 57.5          & 47.3          & 82.9          & 82.1          & 82.2          & 81.0          & 78.6          & 72.2          & 69.4          & 60.2          \\
                                   & 90                                   & 75.4          & 74.5          & 73.3          & 71.9          & 68.8          & 61.3          & 57.1          & 47.0            & 82.6          & 82.0          & 82.0          & 80.4          & 77.6          & 71.9          & 68.5          & 60.4          \\
                                   & 100                                  & 74.8          & 73.6          & 72.5          & 70.9          & 67.6          & 60.5          & 56.6          & 47.1          & 82.2          & 81.2          & 81.1          & 79.4          & 76.8          & 71.5          & 67.9          & 59.9          \\
                                   \hline
\multirow{4}{*}{\texttt{topt-te}} & 20                                   & 74.7          & 74.7          & 74.4          & 74.1          & 72.0            & 67.7          & 64.7          & 53.8          & 80.5          & 80.7          & 80.5          & 80.9          & 79.1          & 77.1          & 74.9          & 67.1          \\
								  & 50                                   & 75.7          & \textbf{75.4}          & \textbf{75.2}          & 74.7 & 72.1          & 64.9          & 60.4          & 49.1          & 82.2          & 81.6          & 82.6          & \textbf{82.4} & 79.9          & 75.0          & 72.0          & 63.0          \\
                                   & 70                                   & 75.7          & 75.3          & 74.4 & 73.3          & 70.0            & 62.3          & 58.4          & 47.4          & 82.9          & 82.3          & \textbf{82.7} & 81.8          & 78.5          & 73.5          & 70.6          & 60.7          \\
                                   & 100                                  & 74.6          & 73.4          & 72.3          & 70.5          & 66.9          & 59.5          & 56.3          & 46.6          & 82.5          & 81.0          & 81.7          & 79.0          & 76.7          & 70.7          & 67.9          & 59.9          \\
                                   \hline
Ours & n/a                                                   & 75.9          & \textbf{75.4}          & \textbf{75.2}          & \textbf{75.2}          & \textbf{74.4}          & \textbf{73.7}          & \textbf{73.1} & \textbf{69.9}          & 82.7          & 81.4          & 81.4          & 81.5          & 79.8          & \textbf{80.0}          & \textbf{80.3} & \textbf{78.6}  \\        
\hline
\end{tabular}
}
\vspace{-1em}
\caption{The MARS test performance (mAP and rank-1 accuracy) of our method learned clip similarity
aggregation and of the top-t\% aggregation baseline. The blocks of rows labeled \texttt{topt-e} and
\texttt{topt-te} show the results of the \texttt{top-t\%-eval} and \texttt{top-t\%-traineval}
variants of the baseline top-$t\%$ clip-similarity aggregation, respectively. \vspace{-1em}}
\label{tab:clip-corruption}
\end{table*}

We compare our method with the \textbf{top-$t\%$ clip-similarity aggregation} (\texttt{top-t\%}) baseline, which is based on \cite{chen2018video}.
It takes $t\%$ of the clip-pairs with highest similarity and averages their similarities to estimate the overall similarity between the two sequences.
By taking only top $t\%$ and not all clip-pairs into account, the resulting similarity becomes more robust and improves re-identification performance \cite{chen2018video}.
In our implementation, we learn a linear layer that projects the $D$-dimensional I3D features to a new $D$-dimensional space.
We define the similarity between two given clips as the cosine similarity between their projected I3D features.
Let $\{{\f'}_q^1, \ldots, {\f'}_q^M\}$ and $\{{\f'}_g^1, \ldots, {\f'}_g^M\}$ be the projected clip features and let $\widehat{P}_t(\x_q, \x_g)$ be the set of $t\%$ clip-pairs with highest similarity.
Then, the \texttt{top-t\%} similarity  between the two sequences is given by,
\begin{align}
\resizebox{0.85\columnwidth}{!}{
$
s_\texttt{top-t\%}(\x_q, \x_g) = \dfrac{1}{\left|\widehat{P}_t(\x_q,
\x_g)\right|}\sum_{\substack{(i,j) \in \\ \widehat{P}_t(\x_q, \x_g)}}\dfrac{{\f'}_q^i\cdot {\f'}_g^j}{\norm{{\f'}_q^i}_2\norm{{\f'}_g^i}_2}. 
$
}
\end{align}

We implement two variants of this method. In the first variant \texttt{top-t\%-eval}, we perform the
top-$t\%$ similarity aggregation only during the evaluation. In the second variant,
\texttt{top-t\%-traineval}, we perform the top-$t\%$ similarity aggregation during the evaluation as
well as during the training. This means that the loss gradients are backpropagated only for the
clips that are included in the top $t\%$ of the clip-pairs.
    
Figure~\ref{fig:top-t-vs-t} shows the test re-identification performance vs $t$ plots for
\texttt{top-t\%-eval} and \texttt{top-t\%-eval} respectively with different values of $M_c^{\max}$.
As expected, the re-identification performance deteriorates as the value of $M_c^{\max}$ is
increased.
We also observe that top-$t\%$ aggregation during training significantly improves the
re-identification performance, especially with the smaller selection rates.

Table~\ref{tab:clip-corruption} shows the re-identification performance of our method and the
baselines on the MARS test set. 
Our method has comparable performance to the top-$t\%$ clip-similarity aggregation when the
corruption rate is low i.e. $M_c^{\max}$ is small. accuracy However, it significantly outperforms
the top-$t\%$ clip-similarity aggregation baseline for higher rates of input corruption, \eg for
$M_c^{\max}=7$ the maximum mAP for the baseline \texttt{topt-e} is $49.3$ (for $t=20\%$), while our
method degrades more gracefully to give $69.6$ mAP. This highlights the advantage of the proposed
learning the clip similarity aggregation.

\subsection{Comparison with the state-of-the-art}

\begin{table}
\resizebox{.9\textwidth}{!}{
\begin{tabular}{l|c|c|c|c}
	\hline
	\textbf{Model}                             & \textbf{mAP}  & \textbf{R1}   & \textbf{R5}   & \textbf{R20}  \\
	\hline
	\rowcolor[HTML]{DCDCDC} 
	RQEN+XQDA+Reranking (2018 \cite{song2018region})      & 71.1 & 77.8 & 88.8 & 94.3 \\
	\rowcolor[HTML]{DCDCDC} 
	TriNet  + Reranking (2017 \cite{hermans2017defense})       & 77.4 & 81.2 & 90.8 & \_   \\
	DuATM (2018 \cite{si2018dual})                     & 67.7 & 81.1 & 92.5 & \_  \\
	MGCAN-Siamese (2018 \cite{song2018mask})           & 71.2 & 77.2 & \_   & \_   \\
	PSE (2018 \cite{saquib2018pose})                  & 56.9 & 72.1 & \_   & \_   \\
	\rowcolor[HTML]{DCDCDC} 
	PSE + ECN (2018 \cite{saquib2018pose})                 & 71.8 & 76.7 & \_   & \_   \\
	RRU + STIM (2018 \cite{liu2018spatial}) *             & 72.7 & 84.4 & 93.2 & 96.3 \\
	Two-Stream M3D (2018 \cite{li2018multi}) *          & 74.1 & 84.4 & 93.8 & 97.7 \\
	PABR (2018 \cite{suh2018part})                     & 75.9 & 84.7 & 94.4 & 97.5 \\
	\rowcolor[HTML]{DCDCDC} 
	PABR + Reranking (2018 \cite{suh2018part})              & 83.9 & 85.1 & 94.2 & 97.4 \\
	CSSA-CSE + Flow (2018 \cite{chen2018video})         & 76.1 & 86.3 & 94.7 & 98.2 \\
	STA (2019 \cite{fu2019sta})                       & 80.8 & 86.3 & 95.7 & 98.1 \\
	\rowcolor[HTML]{DCDCDC} 
	STA + Reranking (2019 \cite{fu2019sta})           & 87.7 & 87.2 & 96.2 & 98.6 \\
	D + GE + $D_G$ (2019 \cite{HuICMR2019}) & 81.8 & 87.3 & 96.0 & 98.1 \\
	\hline
	Ours *                                     & 75.9 & 82.7 & 94.0 & 97.2 \\
	\rowcolor[HTML]{DCDCDC} 
	Ours + Reranking *                         & 83.3 & 83.4 & 93.4 & 97.4 \\
	\hline
\end{tabular}
}
\vspace{-1em}
\caption{Comparison of our model with the state-of-the-art re-identification methods on MARS
dataset. Entries in grey represent the models that use re-ranking. Models marked * use 3D CNNs as
their backbone.} 
\label{tab:sota-mars}
\end{table}

\begin{figure*}
\centering
\vspace{.5em}
\includegraphics[width=\linewidth,trim=0 0 0 60,clip]{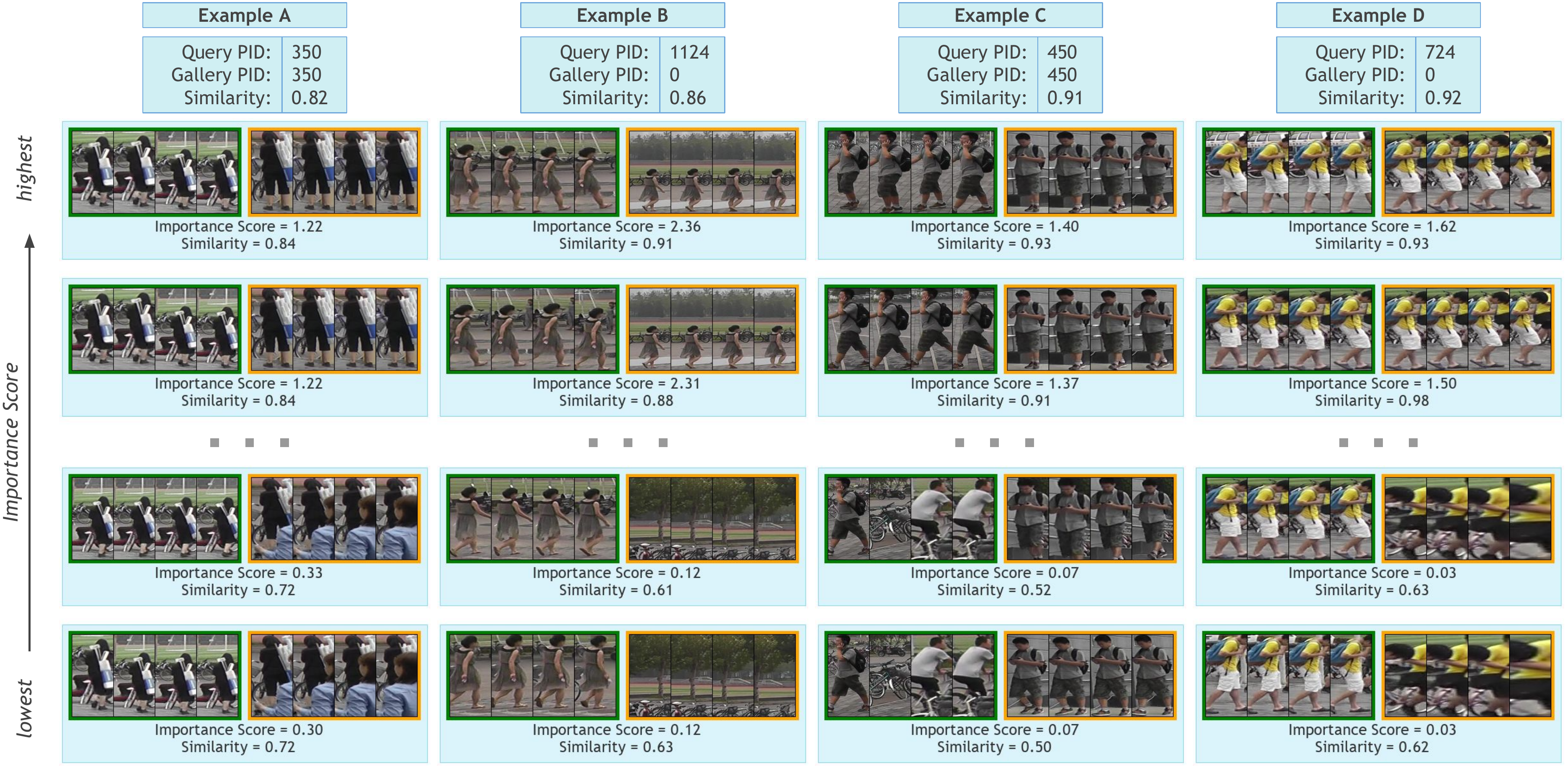}
\vspace{-2em}
\caption{Each column gives an example of different clips from query and gallery videos, where in
each row, left clip is a query clip (green outline) and the right one is a gallery clip (orange
outline). Notice how the method predicts low importance score when the query clip is very different
from the gallery clip, and thus effectively ignores the pair even if the similarity is predicted to
be non trivially high by the feature matching.} 
\label{fig:qual-clip-pairs}
\end{figure*}

\begin{table}
\resizebox{.9\textwidth}{!}{
	\begin{tabular}{l|c|c|c|c}
		\hline
		\textbf{Model}         & \textbf{mAP}  & \textbf{R1}   & \textbf{R5}   & \textbf{R20}  \\
		\hline
		ETAP-Net [Supervised] (2018 \cite{wu2018exploit}) & 78.3 & 83.6 & 94.6 & 97.6 \\
		STA (2019 \cite{fu2019sta})         & 94.9 & 96.2 & 99.3 & 99.6 \\
		R + GE + $D_G$ (2019 \cite{HuICMR2019}) & 94.9 & 95.6 & 99.3 & 99.9 \\
		\hline
		Ours                                 & 88.5 & 89.3 & 98.3 & 99.4 \\
		\hline
	\end{tabular}
}
\vspace{-1em}
	\caption{Comparison of our model with the state-of-the-art methods on DukeMTMC-VideoReID dataset.} 
	\label{tab:sota-duke}
\end{table}

\begin{SCtable}[][t]
\resizebox{.65\columnwidth}{!} {
	\begin{tabular}{l|r|r|r}
		\hline
		\textbf{Model}             & \textbf{R1} & \textbf{R5} & \textbf{R20} \\ \hline
		CNN + XQDA (2016 \cite{zheng2016mars})          & 77.3        & 93.5        & 99.3         \\
		E2E AMOC+EpicFlow (2017 \cite{liu2017video}) & 83.7        & 98.3        & 100.0        \\
		QAN (2017 \cite{liu2017quality})                 & 90.3        & 98.2        & 100.0        \\
		M3D+RAL (2018 \cite{li2018multi})           & 91.0        & \_           & \_            \\
		CSSA-CSE (2018 \cite{chen2018video})            & 88.6        & 99.1        & \_            \\
		Ours                       & 82.9        & 95.8        & 99.1         \\ \hline

		\color{gray}{Two-Stream M3D (2018)}  & \color{gray}{94.4} & \color{gray}{100.0} & \color{gray}{\_}   \\
		\color{gray}{CSSA-CSE + Flow (2018)} & \color{gray}{93.0} & \color{gray}{99.3}  & \color{gray}{100.0}\\ \hline
	\end{tabular}
}
	\caption{Comparison with state-of-the-art re-id methods on PRID2011 dataset.} 
	\label{tab:sota-prid}
\end{SCtable}

In Table~\ref{tab:sota-mars}, we compare our method with the state-of-the-art techniques on MARS dataset.
Our method achieves 75.9\% mAP and 82.7\% Rank-1 accuracy. In terms of mAP, our method is on-par
with all the methods, except for the recently published 
visual distributional representation based method of Hu and Hauptmann \cite{HuICMR2019}, who achieve
an mAP of 81.8\%, which is significantly higher than ours (we discuss below). 
In terms of mAP performance, our method is very close to the part-aligned
bilinear representations (PABR) \cite{suh2018part} and CSSA-CSE + Flow \cite{chen2018video}.
However, the performance of \cite{chen2018video} is much lower than ours when optical flow is not
used (see CSSA-CSE in Table~\ref{tab:sota-mars}).  Among methods that use 3D CNNs as their backbone
(marked * in Table~\ref{tab:sota-mars}), our method achieves the best mAP performance.

Table~\ref{tab:sota-duke} shows the comparison of our method with the state-of-the-art on
DukeMTMC-VideoReID dataset. There are only few works with results on this dataset.We achieve 88.5\%
mAP and 89.3\% Rank-1 accuracy, which is significantly better than the baseline presented in
\cite{wu2018exploit}. However, the performance of Hu and Hauptmann \cite{HuICMR2019} and
\cite{fu2019sta} is better than our method.

Comparing our method to very recent works such as that of Hu and Hauptmann \cite{HuICMR2019}, we
note that their method is significantly more costly than ours in terms of gallery storage
requirements, and uses a CNN networks which is deeper than ours. While we use a 3D CNN with 22
layers, they use an image based DenseNet CNN with 121 layers. They compare the test video with the
gallery videos by estimating the Wasserstein distance between the densities estimated using KDE.
This requires them to use (and save) all the frames to make inference. While in our case, we use a
limited number of clip features ($\sim 8$) per video. While such accurate method achieves higher
performance, it comes at a significant cost.

STA \cite{fu2019sta} is another recent method with state-of-the-art performance.  While STA focuses
on aggregating features effectively from a small set of input frames (4-8 frames), our method is
more focused on predicting the overall similarities between two long sequences while relying on I3D
for clip-level features (a clip is typically 4-16 frames long). Since the video 
benchmarks contain much longer sequences, our method can be used in conjuction
with \cite{fu2019sta} to further boost the performance as it is complimentary to it.

In Table~\ref{tab:sota-prid}, we show results on PRID2011 dataset. Unfortunately being a video based
end-to-end method, our method seems to overfit severely on the dataset. PRID2011 dataset has only
178 videos from training cf.\ 8,298 in MARS. We see that we are still comparable with initial CNN
based methods (eg.\ CNN+XQDA \cite{zheng2016mars}).  The more recent methods seem to utilize optical flow as input, which
could be leading to some regularization by removing the apperance from the videos. 

\subsection{Qualitative Results}
Figure~\ref{fig:qual-clip-pairs} shows four examples of pairs of query-gallery sequences and
the similarity between them as predicted by our method. For each example, we also show two clip
pairs (4 frames each) with the highest importance scores and two with the lowest importance score.
One of the clips in the bottom clip-pair has, from left to right, (i) a significant
amount of occlusion, (ii) no person in the frame, (iii) different persons in different frames due to
a tracking error, and (iv) an improperly cropped person due to poor bounding box estimation.  Our
method learns to correctly identify the clip pairs that are unreliable for estimating the
overall similarity between the two video sequences, and gives them very low importance scores
(bottom two rows). Our method gives an overall high similarity (the heading of each column) to all
the examples shown in Figure~\ref{fig:qual-clip-pairs} by minimizing the effect of bad clip-pairs.
Although MARS dataset considers the gallery sequences in column 2 and 4 as distractors, the high
similarity estimated by our method is reasonable since they contain the same person as in the query
in many of their frames, and are annotation edge cases.

These qualitative results highlight the ability of the proposed method to identify reliable clip
pairs to match, and filter out unreliable ones despite non trivial appearance
similarities estimated by the base network.

%% file: conclusion.tex
\section{Conclusion}

We addressed the video based person re-identification task and, to the best of our knowledge, 
showed that 3D CNNs can be used competitively for the task. We demonstrated better performance with
3D CNN on RGB images only, cf., existing methods which use optical flow channel in addition to RGB
channel. This is indicative of the fact that 3D CNNs are capable of capturing necessary motion cues
relevant for the task of video based person re-identification.

Further, we proposed a novel clip similarity learning method which identifies clip pairs which
are informative for correlating the two clips. While previous methods used ad-hoc methods to obtain
such pairs, we showed that our method is capable of learning to do so. We showed with simulated
partial corruption of input clips, that the proposed method is robust to nuisances which might
occur as a result of motion blur or partial occlusions. We also verified the intuition used to
develop the method qualitatively.

The proposed method can be seen as an approximate discriminative mode matching method. There have
been recent works using deeper CNN models (121 layers cf.\ 22 here) and more accurate distribution
matching which obtain better results than the proposed method, however, they come at a computational
and storage cost. A future work would be systematically find the balance between the two approaches
to obtain the best performance for a given budget.

%% file: appendix.tex
\section{Further implementation details}
\subsection{Details of training of 3D CNN}
For the training of I3D network, we use the AMSGrad optimizer \cite{j.2018on} with $\beta_1=0.9$, $\beta_2=0.999$.
We use a weight decay of $5.0\times 10^{-4}$. In each training iteration, we use a batch of 32 clips belonging to 8 person identities with 4 instances of each identity i.e. $P=8$ and $K=4$.
The RGB input values are scaled and shifted to be in the range $\left[-1.0, 1.0\right]$. For data augmentation, each input clip is first resized up to $144\times 288$ ($H \times W$) and then a random crop of size $128 \times 256$ is taken.
Input clips are also randomly flipped horizontally with a probabiltiy of 0.5.
For training on MARS dataset, we train the network for 1200 epochs with an intial learning rate of $3.0\times 10^{-4}$.
We reduce the learning rate by a factor of 10 after every 400 epochs.
The margin $m$ in the triplet loss expression is set to 0.3.

\subsection{Details of training of Clip-Similarity Aggregation Module}
For the training of Clip-Similarity Aggregation module, we again use the  AMSGrad optimizer with $\beta_1=0.9$, $\beta_2=0.999$ and a weight decay of $5.0\times 10^{-4}$.
We use a batch size of 48 with $P=12$ and $K=4$.
We use the same input transformations and data augmentation techniques as described for the training of the I3D network.
We train the aggregation module for 12 epochs with an initial learning rate of $3.0\times 10^{-5}$.
We reduce the learning rate by a factor of 10 after 8 epochs.
We set margin $m=1$ in the triplet loss.

\section{Further experiments}
\subsection{Ablation experiment for choice of $M_\text{test}$ and $L_\text{test}$}
Tab.~\ref{tab:nclips-vs-clip-length} shows the re-identification performance (mAP) with averaging
I3D features of multiple clips as we vary the number of clips ($M$) and the clip-length
($L$). We can observe that while $L=16$ has a better performance than $L=4, 8$ when using a
single clip $M=1$, it performs lower when number of clips averaged is larger. For $M=8$, $L=4$ and
$L=8$ have similar performances, i.e.\ $74.9$ vs.\ $75.0$. Considering the higher computational cost
with $L=8$, we have used $L=4$, with higher $M$, for the experiments in the paper.

\begin{table}[t!]
	\begin{tabular}{l|lll}
		\hline
		\multirow{2}{*}{\textbf{M}} & \multicolumn{3}{c}{\textbf{mAP}}                                                     \\ \cline{2-4} 
		                                       & \multicolumn{1}{c|}{$\textbf{L}=4$} & \multicolumn{1}{c|}{$8$} & \multicolumn{1}{c}{$16$} \\ \hline
		1                                      & \multicolumn{1}{l|}{63.3}  & \multicolumn{1}{l|}{68.5}  & 69.8                       \\
		2                                      & \multicolumn{1}{l|}{70.1}  & \multicolumn{1}{l|}{73.0}  & 72.9                       \\
		4                                      & \multicolumn{1}{l|}{74.5}  & \multicolumn{1}{l|}{74.9}  & 73.7                       \\
		8                                      & \multicolumn{1}{l|}{74.9}  & \multicolumn{1}{l|}{75.0}  & 73.7                       \\ \hline
	\end{tabular}
    \caption{Reid mAP with averaging I3D features of multiple clips
    ($M$) for different clip-lengths ($L$). 
    The training clip-length and the testing clip-length are
    set to be equal, i.e. $L_\text{test} = L_\text{train}$. The performance reported here is with
    normalization of features.
    }
	\label{tab:nclips-vs-clip-length}
\end{table}

%% file: reid_paper_arxiv.bbl
\begin{thebibliography}{10}\itemsep=-1pt

\bibitem{ahmed2015improved}
E.~Ahmed, M.~Jones, and T.~K. Marks.
\newblock An improved deep learning architecture for person re-identification.
\newblock In {\em Proceedings of the IEEE conference on computer vision and
  pattern recognition}, pages 3908--3916, 2015.

\bibitem{carreira2017quo}
J.~Carreira and A.~Zisserman.
\newblock Quo vadis, action recognition? a new model and the kinetics dataset.
\newblock In {\em proceedings of the IEEE Conference on Computer Vision and
  Pattern Recognition}, pages 6299--6308, 2017.

\bibitem{chen2018video}
D.~Chen, H.~Li, T.~Xiao, S.~Yi, and X.~Wang.
\newblock Video person re-identification with competitive snippet-similarity
  aggregation and co-attentive snippet embedding.
\newblock In {\em Proceedings of the IEEE Conference on Computer Vision and
  Pattern Recognition}, pages 1169--1178, 2018.

\bibitem{cheng2016person}
D.~Cheng, Y.~Gong, S.~Zhou, J.~Wang, and N.~Zheng.
\newblock Person re-identification by multi-channel parts-based cnn with
  improved triplet loss function.
\newblock In {\em Proceedings of the IEEE Conference on Computer Vision and
  Pattern Recognition}, pages 1335--1344, 2016.

\bibitem{dehghan2015gmmcp}
A.~Dehghan, S.~Modiri~Assari, and M.~Shah.
\newblock Gmmcp tracker: Globally optimal generalized maximum multi clique
  problem for multiple object tracking.
\newblock In {\em Proceedings of the IEEE Conference on Computer Vision and
  Pattern Recognition}, pages 4091--4099, 2015.

\bibitem{ding2015deep}
S.~Ding, L.~Lin, G.~Wang, and H.~Chao.
\newblock Deep feature learning with relative distance comparison for person
  re-identification.
\newblock {\em Pattern Recognition}, 48(10):2993--3003, 2015.

\bibitem{farenzena2010person}
M.~Farenzena, L.~Bazzani, A.~Perina, V.~Murino, and M.~Cristani.
\newblock Person re-identification by symmetry-driven accumulation of local
  features.
\newblock In {\em 2010 IEEE Computer Society Conference on Computer Vision and
  Pattern Recognition}, pages 2360--2367. IEEE, 2010.

\bibitem{felzenszwalb2010object}
P.~F. Felzenszwalb, R.~B. Girshick, D.~McAllester, and D.~Ramanan.
\newblock Object detection with discriminatively trained part-based models.
\newblock {\em IEEE transactions on pattern analysis and machine intelligence},
  32(9):1627--1645, 2010.

\bibitem{fu2019sta}
Y.~Fu, X.~Wang, Y.~Wei, and T.~Huang.
\newblock Sta: Spatial-temporal attention for large-scale video-based person
  re-identification.
\newblock In {\em Proceedings of the Association for the Advancement of
  Artificial Intelligence}. 2019.

\bibitem{gray2008viewpoint}
D.~Gray and H.~Tao.
\newblock Viewpoint invariant pedestrian recognition with an ensemble of
  localized features.
\newblock In {\em European conference on computer vision}, pages 262--275.
  Springer, 2008.

\bibitem{hermans2017defense}
A.~Hermans, L.~Beyer, and B.~Leibe.
\newblock In defense of the triplet loss for person re-identification.
\newblock {\em arXiv preprint arXiv:1703.07737}, 2017.

\bibitem{hirzer11}
M.~Hirzer, C.~Beleznai, P.~M. Roth, and H.~Bischof.
\newblock {Person Re-Identification by Descriptive and Discriminative
  Classification}.
\newblock In {\em {Proc. Scandinavian Conference on Image Analysis (SCIA)}},
  2011.

\bibitem{HuICMR2019}
T.-Y. Hu and A.~G. Hauptmann.
\newblock Multi-shot person re-identification through set distance with visual
  distributional representation.
\newblock In {\em Proceedings of the 2019 on International Conference on
  Multimedia Retrieval}, pages 262--270. ACM, 2019.

\bibitem{ioffe2015batch}
S.~Ioffe and C.~Szegedy.
\newblock Batch normalization: Accelerating deep network training by reducing
  internal covariate shift.
\newblock {\em arXiv preprint arXiv:1502.03167}, 2015.

\bibitem{koestinger2012large}
M.~Koestinger, M.~Hirzer, P.~Wohlhart, P.~M. Roth, and H.~Bischof.
\newblock Large scale metric learning from equivalence constraints.
\newblock In {\em 2012 IEEE conference on computer vision and pattern
  recognition}, pages 2288--2295. IEEE, 2012.

\bibitem{kviatkovsky2013color}
I.~Kviatkovsky, A.~Adam, and E.~Rivlin.
\newblock Color invariants for person reidentification.
\newblock {\em IEEE Transactions on Pattern Analysis and Machine Intelligence},
  35(7):1622--1634, 2013.

\bibitem{li2018multi}
J.~Li, S.~Zhang, and T.~Huang.
\newblock Multi-scale 3d convolution network for video based person
  re-identification.
\newblock {\em arXiv preprint arXiv:1811.07468}, 2018.

\bibitem{li2018diversity}
S.~Li, S.~Bak, P.~Carr, and X.~Wang.
\newblock Diversity regularized spatiotemporal attention for video-based person
  re-identification.
\newblock In {\em Proceedings of the IEEE Conference on Computer Vision and
  Pattern Recognition}, pages 369--378, 2018.

\bibitem{li2014deepreid}
W.~Li, R.~Zhao, T.~Xiao, and X.~Wang.
\newblock Deepreid: Deep filter pairing neural network for person
  re-identification.
\newblock In {\em Proceedings of the IEEE Conference on Computer Vision and
  Pattern Recognition}, pages 152--159, 2014.

\bibitem{liao2015efficient}
S.~Liao and S.~Z. Li.
\newblock Efficient psd constrained asymmetric metric learning for person
  re-identification.
\newblock In {\em Proceedings of the IEEE International Conference on Computer
  Vision}, pages 3685--3693, 2015.

\bibitem{liu2017video}
H.~Liu, Z.~Jie, K.~Jayashree, M.~Qi, J.~Jiang, S.~Yan, and J.~Feng.
\newblock Video-based person re-identification with accumulative motion
  context.
\newblock {\em IEEE transactions on circuits and systems for video technology},
  28(10):2788--2802, 2017.

\bibitem{liu2017quality}
Y.~Liu, J.~Yan, and W.~Ouyang.
\newblock Quality aware network for set to set recognition.
\newblock In {\em Proceedings of the IEEE Conference on Computer Vision and
  Pattern Recognition}, pages 5790--5799, 2017.

\bibitem{liu2018spatial}
Y.~Liu, Z.~Yuan, W.~Zhou, and H.~Li.
\newblock Spatial and temporal mutual promotion for video-based person
  re-identification.
\newblock {\em arXiv preprint arXiv:1812.10305}, 2018.

\bibitem{ma2012local}
B.~Ma, Y.~Su, and F.~Jurie.
\newblock Local descriptors encoded by fisher vectors for person
  re-identification.
\newblock In {\em European Conference on Computer Vision}, pages 413--422.
  Springer, 2012.

\bibitem{mclaughlin2016recurrent}
N.~McLaughlin, J.~Martinez~del Rincon, and P.~Miller.
\newblock Recurrent convolutional network for video-based person
  re-identification.
\newblock In {\em Proceedings of the IEEE conference on computer vision and
  pattern recognition}, pages 1325--1334, 2016.

\bibitem{paisitkriangkrai2015learning}
S.~Paisitkriangkrai, C.~Shen, and A.~Van Den~Hengel.
\newblock Learning to rank in person re-identification with metric ensembles.
\newblock In {\em Proceedings of the IEEE Conference on Computer Vision and
  Pattern Recognition}, pages 1846--1855, 2015.

\bibitem{prosser2010person}
B.~J. Prosser, W.-S. Zheng, S.~Gong, T.~Xiang, and Q.~Mary.
\newblock Person re-identification by support vector ranking.
\newblock In {\em BMVC}, volume~2, page~6, 2010.

\bibitem{j.2018on}
S.~J. Reddi, S.~Kale, and S.~Kumar.
\newblock On the convergence of adam and beyond.
\newblock In {\em International Conference on Learning Representations}, 2018.

\bibitem{ristani2016MTMC}
E.~Ristani, F.~Solera, R.~Zou, R.~Cucchiara, and C.~Tomasi.
\newblock Performance measures and a data set for multi-target, multi-camera
  tracking.
\newblock In {\em European Conference on Computer Vision workshop on
  Benchmarking Multi-Target Tracking}, 2016.

\bibitem{saquib2018pose}
M.~Saquib~Sarfraz, A.~Schumann, A.~Eberle, and R.~Stiefelhagen.
\newblock A pose-sensitive embedding for person re-identification with expanded
  cross neighborhood re-ranking.
\newblock In {\em Proceedings of the IEEE Conference on Computer Vision and
  Pattern Recognition}, pages 420--429, 2018.

\bibitem{si2018dual}
J.~Si, H.~Zhang, C.-G. Li, J.~Kuen, X.~Kong, A.~C. Kot, and G.~Wang.
\newblock Dual attention matching network for context-aware feature sequence
  based person re-identification.
\newblock In {\em Proceedings of the IEEE Conference on Computer Vision and
  Pattern Recognition}, pages 5363--5372, 2018.

\bibitem{song2018mask}
C.~Song, Y.~Huang, W.~Ouyang, and L.~Wang.
\newblock Mask-guided contrastive attention model for person re-identification.
\newblock In {\em Proceedings of the IEEE Conference on Computer Vision and
  Pattern Recognition}, pages 1179--1188, 2018.

\bibitem{song2018region}
G.~Song, B.~Leng, Y.~Liu, C.~Hetang, and S.~Cai.
\newblock Region-based quality estimation network for large-scale person
  re-identification.
\newblock In {\em Thirty-Second AAAI Conference on Artificial Intelligence},
  2018.

\bibitem{srivastava2014dropout}
N.~Srivastava, G.~Hinton, A.~Krizhevsky, I.~Sutskever, and R.~Salakhutdinov.
\newblock Dropout: a simple way to prevent neural networks from overfitting.
\newblock {\em The Journal of Machine Learning Research}, 15(1):1929--1958,
  2014.

\bibitem{su2017pose}
C.~Su, J.~Li, S.~Zhang, J.~Xing, W.~Gao, and Q.~Tian.
\newblock Pose-driven deep convolutional model for person re-identification.
\newblock In {\em Proceedings of the IEEE International Conference on Computer
  Vision}, pages 3960--3969, 2017.

\bibitem{suh2018part}
Y.~Suh, J.~Wang, S.~Tang, T.~Mei, and K.~Mu~Lee.
\newblock Part-aligned bilinear representations for person re-identification.
\newblock In {\em Proceedings of the European Conference on Computer Vision
  (ECCV)}, pages 402--419, 2018.

\bibitem{tran2015learning}
D.~Tran, L.~Bourdev, R.~Fergus, L.~Torresani, and M.~Paluri.
\newblock Learning spatiotemporal features with 3d convolutional networks.
\newblock In {\em Proceedings of the IEEE international conference on computer
  vision}, pages 4489--4497, 2015.

\bibitem{wang2014person}
T.~Wang, S.~Gong, X.~Zhu, and S.~Wang.
\newblock Person re-identification by video ranking.
\newblock In {\em European Conference on Computer Vision}, pages 688--703.
  Springer, 2014.

\bibitem{wu2018exploit}
Y.~Wu, Y.~Lin, X.~Dong, Y.~Yan, W.~Ouyang, and Y.~Yang.
\newblock Exploit the unknown gradually: One-shot video-based person
  re-identification by stepwise learning.
\newblock In {\em Proceedings of the IEEE Conference on Computer Vision and
  Pattern Recognition}, pages 5177--5186, 2018.

\bibitem{xiao2016learning}
T.~Xiao, H.~Li, W.~Ouyang, and X.~Wang.
\newblock Learning deep feature representations with domain guided dropout for
  person re-identification.
\newblock In {\em Proceedings of the IEEE conference on computer vision and
  pattern recognition}, pages 1249--1258, 2016.

\bibitem{xiao2017joint}
T.~Xiao, S.~Li, B.~Wang, L.~Lin, and X.~Wang.
\newblock Joint detection and identification feature learning for person
  search.
\newblock In {\em Proceedings of the IEEE Conference on Computer Vision and
  Pattern Recognition}, pages 3415--3424, 2017.

\bibitem{yan2016person}
Y.~Yan, B.~Ni, Z.~Song, C.~Ma, Y.~Yan, and X.~Yang.
\newblock Person re-identification via recurrent feature aggregation.
\newblock In {\em European Conference on Computer Vision}, pages 701--716.
  Springer, 2016.

\bibitem{zhao2017spindle}
H.~Zhao, M.~Tian, S.~Sun, J.~Shao, J.~Yan, S.~Yi, X.~Wang, and X.~Tang.
\newblock Spindle net: Person re-identification with human body region guided
  feature decomposition and fusion.
\newblock In {\em Proceedings of the IEEE Conference on Computer Vision and
  Pattern Recognition}, pages 1077--1085, 2017.

\bibitem{zhao2013unsupervised}
R.~Zhao, W.~Ouyang, and X.~Wang.
\newblock Unsupervised salience learning for person re-identification.
\newblock In {\em Proceedings of the IEEE Conference on Computer Vision and
  Pattern Recognition}, pages 3586--3593, 2013.

\bibitem{zheng2016mars}
L.~Zheng, Z.~Bie, Y.~Sun, J.~Wang, C.~Su, S.~Wang, and Q.~Tian.
\newblock Mars: A video benchmark for large-scale person re-identification.
\newblock In {\em European Conference on Computer Vision}, pages 868--884.
  Springer, 2016.

\bibitem{zheng2017pose}
L.~Zheng, Y.~Huang, H.~Lu, and Y.~Yang.
\newblock Pose invariant embedding for deep person re-identification.
\newblock {\em arXiv preprint arXiv:1701.07732}, 2017.

\bibitem{zheng2011person}
W.-S. Zheng, S.~Gong, and T.~Xiang.
\newblock Person re-identification by probabilistic relative distance
  comparison.
\newblock In {\em CVPR 2011}, pages 649--656. IEEE, 2011.

\bibitem{zhou2017see}
Z.~Zhou, Y.~Huang, W.~Wang, L.~Wang, and T.~Tan.
\newblock See the forest for the trees: Joint spatial and temporal recurrent
  neural networks for video-based person re-identification.
\newblock In {\em Proceedings of the IEEE Conference on Computer Vision and
  Pattern Recognition}, pages 4747--4756, 2017.

\end{thebibliography}
